\documentclass[runningheads]{llncs}

\usepackage[utf8]{inputenc}

\usepackage[utf8]{inputenc}
\usepackage[T1]{fontenc}
\usepackage{newunicodechar}
\newunicodechar{ſ}{\textsl{s}}

\usepackage{graphicx}
\usepackage{subcaption}

\usepackage{amsmath}

\usepackage{wasysym}
\usepackage{arydshln}
\usepackage{url}
\usepackage{xcolor}
\usepackage{booktabs}
\usepackage{multirow}

\begin{document}
\title{Error Patterns in Historical OCR: A Comparative Analysis of TrOCR and a Vision–Language Model}
\titlerunning{Error Patterns in Historical OCR}
\author{Ari Vesalainen\inst{1,2}\orcidID{0009-0005-8064-2082} \and
Eetu M{\"a}kel{\"a}\inst{2}\orcidID{0000-0002-8366-8414} % \and
Laura Ruotsalainen\inst{1}\orcidID{0000-0002-4057-4143}
\and
Mikko Tolonen\inst{2}\orcidID{0000-0003-2892-8911}}
\authorrunning{A. Vesalainen et al.}
\institute{University of Helsinki, Department of Computer Science, Finland \and
University of Helsinki, Department of Digital Humanities, Finland
\email{firstname.lastname@helsinki.fi}
}
\maketitle              % typeset the header of the contribution
\begin{abstract}
Optical Character Recognition (OCR) of eighteenth-century printed texts remains challenging due to degraded print quality, archaic glyphs, and non-standardized orthography. Although transformer-based OCR systems and Vision–Language Models (VLMs) achieve strong aggregate accuracy, metrics such as Character Error Rate (CER) and Word Error Rate (WER) provide limited insight into their reliability for scholarly use. We compare a dedicated OCR transformer (TrOCR) \cite{li2023trocr} and a general-purpose Vision–Language Model (Qwen) \cite{bai2023qwen} on line-level historical English texts using length-weighted accuracy metrics and hypothesis-driven error analysis.

While Qwen achieves lower CER/WER and greater robustness to degraded input, it exhibits selective linguistic regularization and orthographic normalization that may silently alter historically meaningful forms. TrOCR preserves orthographic fidelity more consistently but is more prone to cascading error propagation. Our findings show that architectural inductive biases shape OCR error structure in systematic ways. Models with similar aggregate accuracy can differ substantially in error locality, detectability, and downstream scholarly risk, underscoring the need for architecture-aware evaluation in historical digitization workflows.

\end{abstract}
\section{Introduction}

Accurate OCR for documents that fall outside modern training distributions remains a major challenge in document analysis \cite{nguyen2021survey}. Despite substantial progress in transformer-based recognizers and end-to-end neural OCR systems, aggregate accuracy metrics alone provide an incomplete account of reliability in real-world digitization pipelines. OCR errors propagate into downstream processes such as corpus construction, linguistic analysis, named-entity recognition, and historical interpretation, where even small deviations can affect meaning, searchability, or evidential reliability \cite{nguyen2021survey}. In such contexts, understanding the structure and origin of errors is as important as measuring their frequency.

Historical and non-standard document collections introduce severe domain shift. Eighteenth-century printed materials exhibit extended glyph inventories (e.g., long-s, ligatures), non-standardized orthography, heterogeneous typefaces, and degradation effects such as ink spread, bleed-through, and low contrast. These conditions give rise not only to elevated error rates, but to systematic error patterns, including visually plausible substitutions, implicit orthographic normalization, and incorrect word segmentation. Such patterns reflect architectural inductive biases and interact differently with downstream correction and analysis workflows.

Recent OCR research has focused primarily on improving aggregate accuracy through architectural advances and larger training corpora. OCR-native encoder--decoder models such as TrOCR emphasize visually grounded character-level alignment \cite{li2023trocr}, while Vision--Language Models (VLMs) formulate transcription as conditional text generation and incorporate strong linguistic priors \cite{greif2025multimodal}. Although these approaches may achieve comparable Character Error Rate (CER) or Word Error Rate (WER) on challenging material, similar scores do not imply similar behavior. The practical impact of OCR output depends on whether errors are local or cascading, visually transparent or linguistically plausible, and whether they preserve or silently alter historical signal.

We therefore analyze OCR models as architectural components embedded in digitization pipelines. Rather than ranking systems along a single performance axis, we examine how different architectural principles shape characteristic error behaviors under identical segmentation and input conditions. By restricting inputs to pre-segmented line images, we isolate recognition dynamics and focus on error structure rather than upstream preprocessing artifacts.

This study uses eighteenth-century European printed texts as a stress test for architectural error behavior under realistic historical conditions. We contrast a visually grounded OCR-native model (TrOCR \cite{li2023trocr}) with a general-purpose Vision--Language Model (Qwen \cite{bai2023qwen}). Our goal is not to identify a universal winner, but to characterize how architectural differences manifest in visually driven confusions, linguistic substitutions, normalization effects, and error propagation.

Our analysis shows that models with comparable CER/WER can generate qualitatively distinct error profiles. Visually grounded architectures tend to produce locally faithful but potentially cascading errors, particularly on long or degraded lines, whereas Vision--Language Models favor bounded and linguistically plausible outputs that may include silent normalization. These differences have direct implications for downstream correction effort, search robustness, and scholarly interpretation.

This paper makes the following contributions:
\begin{itemize}
\item We provide a comparative characterization of architectural error modes in historical OCR, demonstrating how similar aggregate accuracy can mask structurally different recognition behaviors.
\item We develop a hypothesis-driven framework linking architectural properties to observable error patterns, including visual confusions, linguistic substitutions, normalization tendencies, and propagation dynamics.
\item We propose an evaluation perspective that complements aggregate accuracy metrics with error-structure analyses relevant to correction effort, risk, and downstream Digital Humanities applications.
\end{itemize}

Section~2 reviews related work. Section~3 describes the datasets, models, and experimental setup. Section~4 presents the empirical error analyses. Section~5 discusses implications for pipeline design and model selection. Section~6 concludes with limitations and future directions.

\section{Related Work}

OCR for historical and early printed texts is widely recognized as more challenging than OCR for modern documents. Eighteenth-century materials exhibit non-standardized orthography, archaic glyphs such as the long-s, frequent ligatures, heterogeneous typefaces, and physical degradation, all of which degrade recognition performance and limit the effectiveness of off-the-shelf OCR systems \cite{reul2019ocr,springmann2018groundtruth}. While neural network–based approaches combined with domain-specific training and post-correction have improved accuracy, recognition quality remains uneven across fonts, scripts, languages, and scan conditions \cite{springmann2018groundtruth,neudecker2019ocrd,drobac2020postcorrection}. Importantly, even low character error rates on curated datasets do not eliminate systematic errors that can materially affect downstream scholarly analysis \cite{hill2019quantifying}.

As a result, increasing attention has shifted from aggregate accuracy to the structure of OCR errors. Prior work demonstrates that OCR errors are not random noise but follow recurring patterns, including character substitutions, insertions, deletions, word-boundary errors, and positional effects \cite{wick2018comparison,nguyen2019error}. Such error taxonomies are particularly relevant for historical texts, where error distributions are shaped by typography, print practices, and script conventions rather than stochastic variation \cite{dhondt2016lowresource}. Detailed error analyses have supported targeted post-correction strategies and revealed persistent failure modes tied to specific glyphs, fonts, and historical periods \cite{nguyen2019error,drobac2020postcorrection}.

Transformer-based OCR architectures represent a major shift in OCR system design. By combining a visual encoder with an autoregressive text decoder trained on aligned image--text pairs, these models achieve strong performance on modern and historical printed materials when fine-tuned on domain-specific data \cite{li2023trocr,strobel2023adaptability,cheema2024multilingual}. However, prior studies demonstrate that transformer-based OCR models remain sensitive to segmentation quality, unseen fonts and scripts, and domain-specific degradation, particularly in historical and early printed materials \cite{mostafa2021ocformer,strobel2023adaptability}.

More recently, Vision--Language Models (VLMs) have been applied to OCR and document transcription tasks, formulating transcription as conditional text generation within a unified multimodal framework. Large-scale evaluations report strong performance on multilingual, noisy, and complex-layout OCR tasks, sometimes without explicit character-level supervision \cite{ye2023ureader,liu2024ocr}.  Applied to historical documents, VLMs show robustness to noise and layout variability but also exhibit characteristic behaviors such as orthographic normalization, modernization of spelling, and hallucinated content in visually ambiguous regions \cite{vogler2021lacuna,greif2025multimodal,liu2024ocr}. While these behaviors may improve surface readability, they raise concerns for diplomatically faithful transcription and historical analysis.

These architectural differences have direct implications for OCR error behavior. OCR-native transformer models are trained with loss functions that penalize character-level deviations, encouraging fidelity to the visual signal. Vision--Language Models, by contrast, tolerate minor visual inconsistencies when semantic plausibility is preserved, reflecting their generative training objectives \cite{hill2019quantifying}. Consequently, models with similar CER or WER may exhibit qualitatively different error profiles with distinct scholarly consequences \cite{strobel2023adaptability}.

Despite extensive work on OCR accuracy and error categorization, few studies systematically compare how different model architectures shape error behavior in historical OCR. In particular, the interaction between architectural bias, linguistic priors, and downstream scholarly risk remains underexplored for eighteenth-century printed materials characterized by strong orthographic variability and print degradation \cite{hill2019quantifying}.

Finally, while CER and WER remain standard evaluation metrics, their limitations for historical texts are well documented. These metrics penalize historically valid spelling variants, obscure normalization effects, and fail to capture layout- and segmentation-induced errors \cite{hill2019quantifying,neudecker2019ocrd,dias2023layout}. Recent work therefore advocates richer evaluation frameworks that combine quantitative error typologies with qualitative and task-sensitive analysis, better aligning OCR evaluation with scholarly practice \cite{nguyen2021frameworks,greif2025multimodal}.

\section{Methodology}

\subsection{Training Data}

\sloppy {To adapt OCR models to the typographic and material characteristics of \allowbreak eighteenth-century printed texts, we constructed a custom line-level training corpus that combines manually annotated, semi-automatically generated, and synthetic data. The corpus is designed to balance transcription fidelity, historical and typographic diversity, and scalability, while preserving orthographic and graphemic variation characteristic of early print. All data are prepared at line level to match the experimental setup and evaluation protocol.}

The training data consist of the following components:
\begin{itemize}
    \item \textbf{Manually annotated data} (approximately 62{,}000 line images), cropped from color scans of eighteenth-century books, primarily from McGill University Library collections. These include running text as well as common non-body elements such as marginalia, footnotes, page numbers, catchwords, and signature marks. Ground truth transcriptions were produced by manual correction of initial OCR outputs, yielding high-fidelity labels that preserve original orthography and visual artifacts.
    \item \sloppy {\textbf{Semi-automated data} (approximately 68{,}000 line images), derived from ECCO page images aligned with corresponding transcriptions from the ECCO-TCP corpus \cite{Gregg_2021}.\footnote{The Eighteenth Century Collections Online (ECCO) Text Creation Partnership (TCP) project.} Line-level silver-standard pairs were extracted using edit-distance–based alignment and filtered to remove structurally unreliable content, increasing typographic and bibliographic diversity while remaining grounded in authentic historical sources.}
    \item \textbf{Synthetic data} (approximately 65{,}000 line images), generated using historically informed serif typefaces and document backgrounds sampled from real scans. Synthetic lines are used to increase exposure to rare glyphs, spelling variants, and degradation patterns that occur infrequently in manually annotated data.
\end{itemize}

In total, the training corpus comprises approximately 195{,}000 line images. The combination of manual, semi-automatic, and synthetic data enables robust supervision across a wide range of eighteenth-century typographic conditions. The dataset is publicly available via Zenodo.\footnote{\url{https://zenodo.org/records/18597235}}

\subsection{Models and Fine-Tuning}

We compare two OCR approaches with distinct architectural assumptions: an OCR-native transformer (TrOCR) and a general-purpose Vision--Language Model (Qwen).

\emph{TrOCR.} We fine-tuned a TrOCR model initialized from the \texttt{microsoft/trocr-\allowbreak large-\allowbreak printed} checkpoint using line-level image--text pairs. Training followed the standard TrOCR pipeline, with visual augmentations applied to improve robustness to historical scanning artifacts. All experiments reported in this paper use the TrOCR-large model in order to avoid confounding architectural effects with model capacity.

\emph{Qwen.} Qwen2.5-VL-7B-Instruct is a large Vision--Language Model pretrained via multimodal and instruction tuning, with OCR learned implicitly rather than through explicit character-level supervision. We fine-tuned Qwen for line-level transcription by framing OCR as an instruction-following task applied to line images, using parameter-efficient fine-tuning while freezing the vision encoder.

The compared models differ substantially in scale and pretraining regime. While Vision--Language Models such as Qwen incorporate far more parameters and broader pretraining than OCR-native transformers, the goal of this study is not to attribute performance differences to model size. Instead, we focus on how architectural inductive biases shape OCR error behavior when models operate on identical, controlled inputs.

\subsection{Line-Level Inputs for Vision--Language OCR}

Although Qwen is capable of processing full-page images, both models are restricted to line-level inputs in order to ensure comparability and preserve editorially significant structure. Line-level processing prevents cross-line interactions such as implicit merging of hyphenated line breaks and constrains recognition to the visual evidence present within each line. This design choice enables controlled analysis of recognition behavior and error patterns while maintaining fidelity to transcription practices required for critical editions.

\subsection{Evaluation Datasets}

All evaluations are conducted on datasets that are strictly disjoint from the training data, with disjointness enforced at the book level to prevent typographic or lexical leakage. A detailed breakdown of training and evaluation splits is publicly available to support verification and reproducibility.\footnote{\url{https://github.com/tbd}}
% https://github.com/HISCOM/error_patterns_in_historical_ocr
\emph{English Historical Dataset.}
The primary evaluation corpus consists of approximately 200 pages (about 10{,}000 line images) of English historical printed texts from the early eighteenth century onward. The material includes a range of genres and typographic conditions, with source scans drawn from both binarized black-and-white ECCO images and higher-quality grayscale or color images from the Internet Archive. This dataset supports controlled analysis of robustness to visual degradation, scan modality, and historically non-standardized orthography, and forms the basis for the detailed error analyses reported in Section~4.

Although both models are trained on comparable line-level data, their differing architectural inductive biases motivate the hypothesis-driven error analyses presented in the following section.

\subsection{Transcription, Normalization, and Operational Setup}
\label{sec:transcription_normalization}

All models in this study are trained and evaluated against a \emph{diplomatic transcription target with minimal, technical normalization}. The transcription policy is designed to preserve graphemic and typographic characteristics specific to eighteenth-century print, while eliminating only those representational variations that are unrelated to visual character recognition and would otherwise confound training and evaluation. The policy is model-agnostic and applied uniformly across ground truth preparation, inference output, and evaluation, with no model-specific normalization, correction, or post-hoc adjustment.

The target transcription preserves original printed forms without modernization or editorial intervention. Historically meaningful graphemes, including long~\emph{s} and typographic ligatures, are retained as they appear in the source material. Original diacritics, punctuation, and capitalization are preserved, and no spelling normalization, abbreviation expansion, or lexical regularization is introduced. This ensures fidelity to the visual and orthographic conventions of eighteenth-century print.

A limited and deterministic form of normalization is applied solely to stabilize encoding-level variation that is not visually or linguistically informative. Specifically, consecutive whitespace is collapsed, leading and trailing whitespace is removed, and Unicode dash or hyphen variants are mapped to a single ASCII hyphen, with repeated hyphens collapsed. These operations affect only spacing and punctuation encoding and do not alter underlying graphemic content. The same normalization policy is applied identically to reference transcriptions and model outputs prior to evaluation, ensuring that reported character error rates (CER) reflect grapheme-level recognition errors rather than representational artifacts.

\sloppy{All experiments operate on pre-segmented line images. No binarization, deskewing, or contrast enhancement is applied, in order to preserve typographic variation characteristic of historical print. Images are provided to each model using its standard preprocessing pipeline. Due to architectural differences, image resizing is handled differently: TrOCR requires explicit resizing to a fixed input resolution defined by its vision encoder, which may distort aspect ratios for long or narrow lines, whereas Qwen accepts variable-resolution images and performs internal rescaling while preserving aspect ratio. These differences are inherent to the respective architectures and are not compensated for during preprocessing.}

Decoding is performed deterministically for all models using fixed generation parameters. Apart from resizing behavior imposed by the models themselves, all other operational settings are kept constant across experiments, and no additional post-processing is applied beyond the transcription and minimal normalization policy described above.

\section{Experimental Results}

This section presents a comparative evaluation of TrOCR and Qwen on two historical OCR test sets. We first report aggregate accuracy using standard OCR metrics and then conduct a detailed error analysis following the framework of Nguyen et al.\cite{nguyen2019error}, enabling systematic comparison beyond CER and WER.

\subsection{Quantitative Performance}

\paragraph{Evaluation Metrics.}
CER and WER are computed using standard Levenshtein-based definitions at the character and word levels, respectively.
We report length-weighted variants of CER and WER. For brevity, all subsequent references to CER/WER denote the length-weighted versions unless explicitly stated otherwise. Confidence intervals are estimated using paired nonparametric bootstrap resampling over line instances, ensuring that identical inputs are compared across models.

Table~\ref{tab:global_cer_wer} reports CER and WER for Qwen and TrOCR, reported separately by scan modality for test dataset (color vs.\ black-and-white). Across all evaluation settings, Qwen achieves lower error rates than TrOCR.

For test material, color scans yield substantially lower error rates than binarized black-and-white images for both models, reflecting the higher visual quality of color scans compared to ECCO-derived inputs. This effect is particularly pronounced for TrOCR, whose CER increases from 0.82\% on color images to 1.46\% on black-and-white images, indicating greater sensitivity to image degradation. Qwen shows a smaller but consistent degradation under binarization, suggesting increased robustness to reduced visual fidelity.

\begin{table}[t]
\centering
\caption{Global OCR accuracy statistics (mean $\pm$ 95\% CI) for Qwen and TrOCR across all images and by image type.}
\label{tab:global_cer_wer}
\begin{tabular}{llcc}
\hline
\textbf{Subset} & \textbf{Model} & \textbf{CER (\%)} & \textbf{WER (\%)} \\
\hline
\multirow{2}{*}{All images} 
  & Qwen  & $0.5986 \pm 0.0428$ & $2.4771 \pm 0.1326$ \\
  & TrOCR & $1.0517 \pm 0.0677$ & $3.8909 \pm 0.1632$ \\
\hline
\multirow{2}{*}{Coloured images} 
  & Qwen  & $0.5001 \pm 0.0465$ & $2.2540 \pm 0.1681$ \\
  & TrOCR & $0.8157 \pm 0.0602$ & $3.3974 \pm 0.1933$ \\
\hline
\multirow{2}{*}{Black \& white images} 
  & Qwen  & $0.7679 \pm 0.0828$ & $2.8654 \pm 0.2295$ \\
  & TrOCR & $1.4578 \pm 0.1711$ & $4.7496 \pm 0.3349$ \\
\hline
\end{tabular}
\end{table}

Table~\ref{tab:cer-wer-by-line-length} summarizes OCR accuracy by line-length category. For both models, error rates decrease as line length increases, with short lines showing the highest CER and WER. This effect is particularly strong for TrOCR, indicating that limited visual and linguistic context makes short segments more challenging.

Qwen outperforms TrOCR across all length categories, with the largest relative gap observed for long lines (CER of 0.54\% vs.\ 0.98\%). As line length increases, error rates decrease sharply for both models and the performance gap narrows, suggesting that additional textual context benefits both visually grounded and language-aware decoding.

\begin{table}[t]
\centering
\caption{Global OCR accuracy statistics (mean $\pm$ 95\% CI) for Qwen and TrOCR by line-length category.}
\label{tab:cer-wer-by-line-length}
\begin{tabular}{llcc}
\toprule
\textbf{Line length} & \textbf{Model} & \textbf{CER (\%)} & \textbf{WER (\%)} \\
\midrule
\multirow{2}{*}{Short}
  & Qwen  & $3.9263 \pm 1.5643$ & $9.1938 \pm 2.6522$ \\
  & TrOCR & $5.7883 \pm 1.3407$ & $15.2758 \pm 2.5528$ \\
\midrule
\multirow{2}{*}{Medium}
  & Qwen  & $1.5636 \pm 0.3778$ & $5.7182 \pm 1.1507$ \\
  & TrOCR & $2.2191 \pm 0.5248$ & $7.5938 \pm 1.3338$ \\
\midrule
\multirow{2}{*}{Long}
  & Qwen  & $0.5422 \pm 0.0402$ & $2.2875 \pm 0.1266$ \\
  & TrOCR & $0.9766 \pm 0.0711$ & $3.6142 \pm 0.1680$ \\
\bottomrule
\end{tabular}
\end{table}

As expected for a large Vision–Language Model, Qwen achieves substantially lower aggregate error rates. While these aggregate metrics indicate a clear performance advantage, they do not reveal the nature or causes of remaining errors. We therefore complement accuracy evaluation with detailed error analysis.

\subsection{Error taxonomy and proxy definitions}
\label{sec:error_taxonomy}

\sloppy{Error analysis follows the proxy-based OCR error taxonomy proposed by Nguyen et~al.~\cite{nguyen2019error}, adapted to the characteristics of eighteenth-century printed text. All error proxies are computed after applying the transcription and normalization policy described in Section~\ref{sec:transcription_normalization}.}

We distinguish between \emph{real-word} and \emph{non-word} errors based on membership in a historically informed lexicon. A real word is defined as a token attested in eighteenth-century sources, as represented by a lexicon constructed from ECCO-TCP transcriptions and additional historical English corpora, including preserved graphemic variants such as long-s and typographic ligatures. Tokens not present in this lexicon after normalization are classified as non-words. No modern spell-checking, lemmatization, or language-model–based correction is applied.

Character-level errors are identified through standard alignment and include substitutions, insertions, and deletions. As these categories are well established in OCR evaluation, we treat them as a unified class of character errors and focus instead on their interaction with higher-level error proxies. In addition, we identify \emph{boundary errors}, corresponding to incorrect token splitting or merging caused by misplaced or missing whitespace.

Boundary errors are detected using alignment-based heuristics that compare token boundaries before and after normalization, allowing merged and split tokens to be identified deterministically. Finally, we track systematic \emph{glyph confusions}, such as substitutions between historically distinct but visually similar forms (e.g.\ long-s versus modern s, or ligatures versus decomposed sequences), as a separate diagnostic category.

All lexicon construction steps, proxy computations, error classification rules, and the full fine-tuning and inference pipelines are deterministic and released as open-source code to ensure reproducibility.\footnote{\url{https://github.com/tbd}}
% https://github.com/HISCOM/error_patterns_in_historical_ocr
\begin{table}[t]
\centering
\caption{Definition of error proxies used in evaluation.}
\label{tab:error_proxies}
\begin{tabular}{p{3.5cm} p{7.5cm}}
\hline
\textbf{Proxy} & \textbf{Operational definition} \\
\hline
Real-word error & Token differs from reference but remains attested in historical lexicon \\
Non-word error & Token differs from reference and is unattested in historical lexicon \\
Boundary error & Token split or merge due to incorrect whitespace placement \\
Glyph confusion & Systematic substitution between historically distinct glyph forms \\
\hline
\end{tabular}
\end{table}

\subsection{Detailed Error Analysis}

\begin{figure}[!t]
    \centering

    % Row 1
    \begin{subfigure}[t]{0.49\linewidth}
        \centering
        \includegraphics[width=\linewidth]{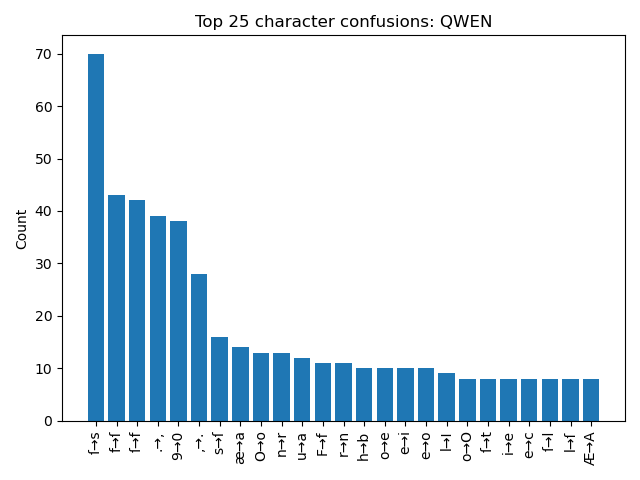}
        \caption{Top-25 character confusions for Qwen }
        \label{fig:confusion_qwen}
    \end{subfigure}\hfill
    \begin{subfigure}[t]{0.49\linewidth}
        \centering
        \includegraphics[width=\linewidth]{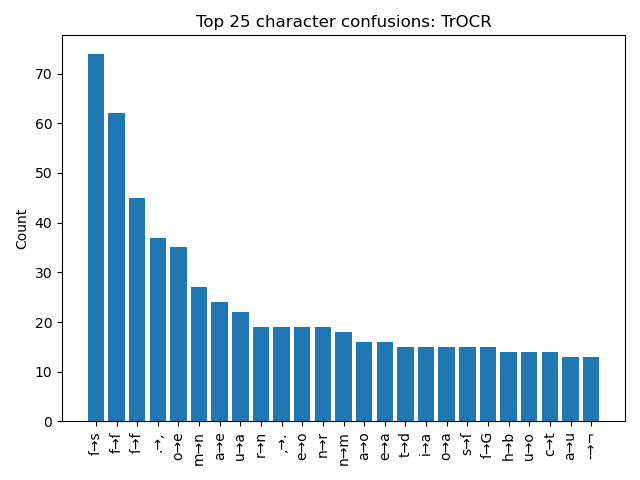}
        \caption{Top-25 chracter confusions for TrOCR}
        \label{fig:confusion_trocr}
    \end{subfigure}

    \vspace{0.5em}

    % Row 2
    \begin{subfigure}[t]{0.49\linewidth}
        \centering
        \includegraphics[width=\linewidth]{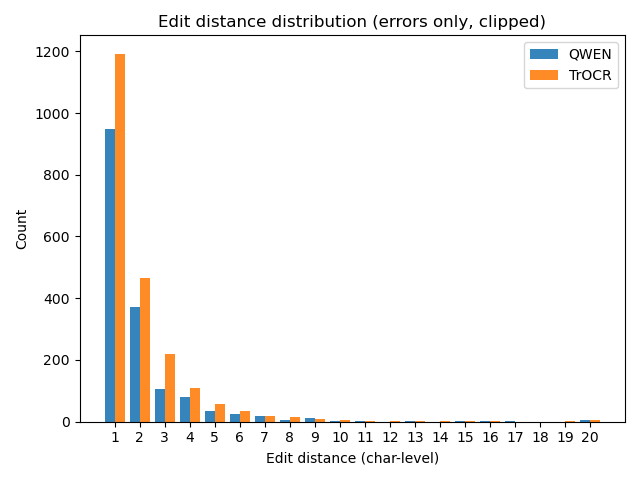}
        \caption{Edit distance distribution}
        \label{fig:edit_distance}
    \end{subfigure}\hfill
    \begin{subfigure}[t]{0.49\linewidth}
        \centering
        \includegraphics[width=\linewidth]{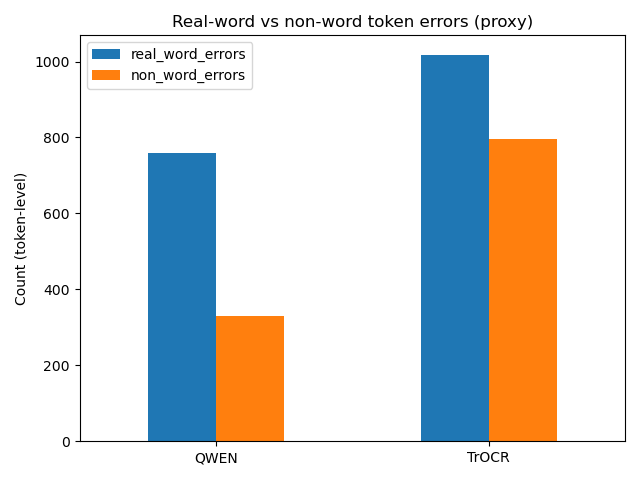}
        \caption{Share of Words and Non-words}
        \label{fig:word_vs_non-word}
    \end{subfigure}

    \vspace{0.5em}

    % Row 3
    \begin{subfigure}[t]{0.39\linewidth}
        \centering
        \includegraphics[width=\linewidth]{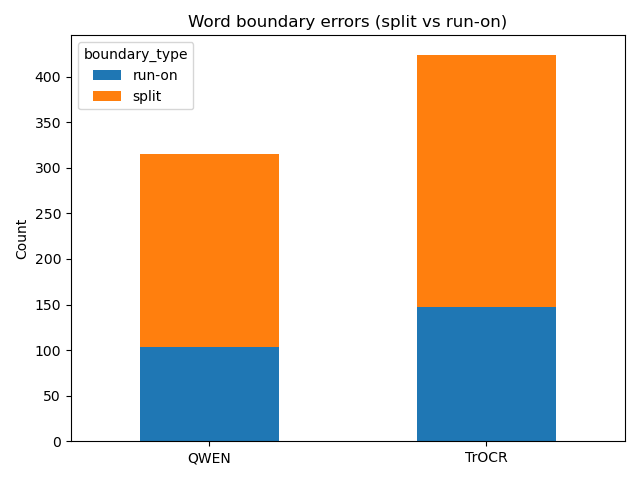}
        \caption{Word boundary errors}
        \label{fig:boundary_errors}
    \end{subfigure}\hfill
    \begin{subfigure}[t]{0.59\linewidth}
        \centering
        \includegraphics[width=\linewidth]{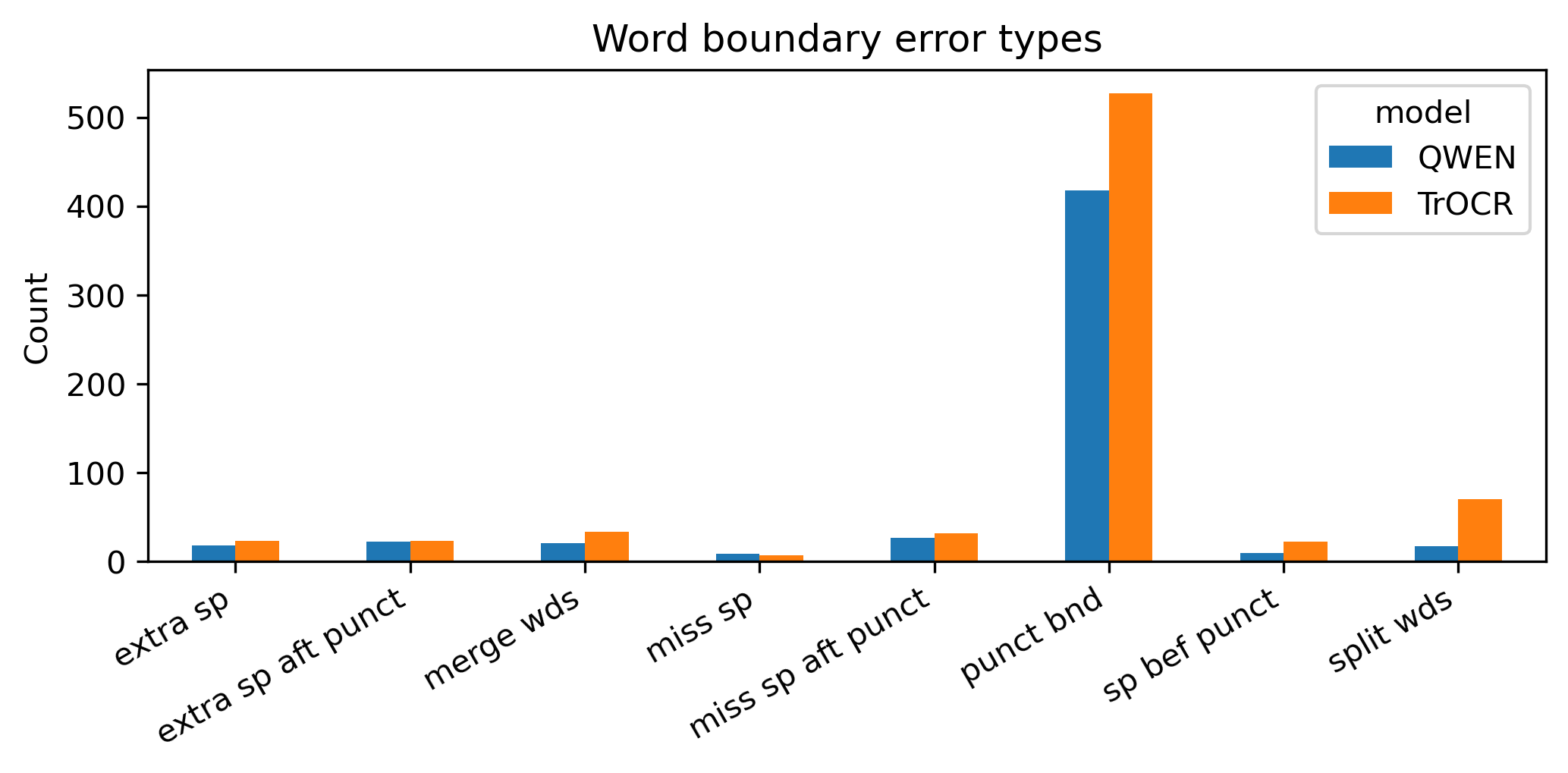}
        \caption{Types of word boundary errors}
        \label{fig:boundary_error_trypes}
    \end{subfigure}

    \caption{Illustrations of error analysis}
    \label{fig:grid3x2}
\end{figure}

Building on the proxy definitions introduced in Section~\ref{sec:error_taxonomy}, we analyze OCR error behavior across multiple orthogonal dimensions that characterize error structure rather than aggregate accuracy (Figure~\ref{fig:grid3x2}).

\emph{Edit Operations and Length Effects.}  

Both models exhibit systematic confusions involving visually similar glyphs, such as \emph{\longs~$\rightarrow$~s}, \emph{\longs~$\rightarrow$~f}, and punctuation substitutions, reflecting typographic ambiguity and print degradation (Figures~\ref{fig:confusion_qwen} and~\ref{fig:confusion_trocr}). While the same confusion types occur for both architectures, their relative frequencies differ, indicating distinct inductive biases. Error distributions are dominated by low edit-distance operations (Figure~\ref{fig:edit_distance}): 82.0\% of Qwen’s errors and 77.5\% of TrOCR’s errors have edit-distance 1 or 2. This confirms that most deviations are highly localized, with Qwen producing a slightly larger share of minimal edits. Recognition performance also varies systematically with line length: short segments exhibit the highest error rates, while longer segments achieve lower CER and WER for both models, with TrOCR showing a steeper degradation as sequence length increases.

\emph{Erroneous Character Positions.}  
Errors are not uniformly distributed along the text line. Across both models, short and medium-length segments exhibit a tendency toward line-initial errors, while longer sequences show a more even distribution with increased mid-line error concentration. These positional patterns are consistent with autoregressive decoding dynamics and affect downstream alignment and correction behavior.

\emph{Real-Word vs.\ Non-Word Errors.}  
Figure~\ref{fig:word_vs_non-word} shows that a substantial fraction of OCR errors produced by both models result in valid dictionary entries. As a consequence, many errors cannot be detected using lexical filtering or dictionary-based post-correction alone, highlighting the limits of surface-level validation for historical OCR.

\emph{Word-Boundary Errors.}  
Both architectures produce word-boundary errors caused by erroneous insertion or deletion of whitespace, leading to incorrect word splits and run-on forms (Figure~\ref{fig:boundary_errors}). Boundary errors are dominated by punctuation-related cases, reflecting the difficulty of handling punctuation and surrounding whitespace in degraded historical material (Figure~\ref{fig:boundary_error_trypes}). While Qwen produces fewer boundary errors overall, TrOCR shows higher sensitivity to segmentation ambiguity around punctuation.

Across all evaluation settings, Qwen achieves lower aggregate error rates than TrOCR. Beyond these metrics, the analyses reveal systematic, model-specific differences in error structure. TrOCR exhibits stronger error accumulation with increasing sequence length and greater sensitivity to segmentation ambiguity, whereas Qwen produces more localized errors and a higher share of real-word substitutions. These findings underscore that aggregate accuracy alone does not capture the structure or impact of OCR errors, motivating the architectural synthesis in Section~5.

\section{Error Analysis and Architectural Impact}

The results in Section~4 show that differences between TrOCR and Qwen extend beyond aggregate accuracy and manifest in systematic differences in error behavior, robustness, and scalability. These differences cannot be explained by training data size or evaluation conditions alone, indicating that architectural design choices play a central role in OCR performance on historical material. This section therefore examines how visual representations, decoding strategies, and language integration mechanisms shape the observed error patterns.

We analyze TrOCR and Qwen on noisy eighteenth-century printed texts. Rather than treating performance differences as purely empirical outcomes, we formulate four hypotheses linking architectural properties to characteristic error behaviors. Together, these hypotheses address visual grounding, linguistic regularization, orthographic normalization, and error dynamics, providing a structured framework for interpreting the empirical findings.

\emph{Visual Grounding Hypothesis (H1).}  
Hypothesis~H1 posits that an OCR-native architecture such as TrOCR exhibits more visually grounded error behavior than a vision--language Model, producing a higher proportion of character-level substitutions driven by visual ambiguity.

The results provide consistent support for this hypothesis. TrOCR produces more visually motivated substitutions than Qwen, particularly for historically ambiguous glyphs such as long-s and minim-based sequences. These errors tend to remain close to observed glyph shapes, even when the resulting output is lexically invalid, reflecting TrOCR’s emphasis on faithful visual transcription through explicit character-level supervision. By contrast, Qwen resolves many visually uncertain cases using linguistic priors, favoring lexically plausible outputs over strict visual fidelity. Representative examples are shown in Table~\ref{tab:H1}, and bootstrap analysis confirms a stable directional bias toward visually grounded substitutions in TrOCR.

\begin{table}[t] 
\centering \caption{Representative examples of visual grounding} \label{tab:H1} \begin{tabular}{p{3.8cm}p{3.8cm}p{3.8cm}} \hline 
\textbf{Ground Truth} & \textbf{TrOCR Output} & \textbf{QWEN Output} \\ 
\hline 
\texttt{pre\longs ent conduct of affairs, and \longs uch apprehen\longs ions were entertained of futurity,} & \texttt{pre\longs ent conduct of affairs, and \longs uch apprehen\longs ions were entertained of \emph{\longs uturity},} & \texttt{pre\longs ent conduct of affairs, and \longs uch apprehen\longs ions were entertained of futurity,} \\ 
\hdashline \texttt{\longs ummon it, 322. One \longs ummoned under the influence of the covenant} & \texttt{\longs ummon it, 322. One \emph{\longs ummened} under the influence of the covenant-} & 
\texttt{\longs ummon it, 322. One \longs ummoned under the influence of the covenant-} \\ 
\hline 
\end{tabular} 
\end{table}

\emph{Linguistic Regularization Hypothesis (H2).}  
Hypothesis~H2 posits that a vision--language Model with a strong generative language component will resolve visual uncertainty through linguistic regularization, producing lexically plausible substitutions.

Aggregate real-word ratios alone do not clearly separate the models, as both produce substantial proportions of real-word errors. However, qualitative analysis reveals systematic differences in how such errors arise. Qwen more frequently substitutes visually uncertain or historically specific forms with semantically and syntactically plausible alternatives, reflecting context-sensitive generative decoding. TrOCR, in contrast, produces real-word errors less systematically and more often alongside non-word outputs, indicating weaker lexical constraint. As illustrated in Table~\ref{tab:h2_examples}, linguistic regularization in Vision--Language Models manifests selectively through plausible substitutions rather than as a uniform shift in error composition, increasing the risk of silent semantic distortion in downstream analysis.

\begin{table}[t]
\centering
\caption{Examples of linguistic  regularization errors.}
\label{tab:h2_examples}
\begin{tabular}{p{2.5cm} p{3.2cm} p{7cm}}
\hline
\textbf{GT word} & \textbf{OCR word} & \textbf{Context (GT $\rightarrow$ OCR)} \\
\hline
tarded & seemed & tarded not the execution $\rightarrow$ seemed not the execution \\
rigorouſly & merely & rigorouſly enforced $\rightarrow$ merely enforced \\
Artay & free & Artay, commissions of $\rightarrow$ free commissions of \\
Groans & Gentleman & Groans of the Britains $\rightarrow$ Gentleman of the Britains \\
ward & used & ward to assert $\rightarrow$ used to assert \\
\hline
\end{tabular}
\end{table}

\emph{Orthographic Normalization Hypothesis (H3).}  
Hypothesis~H3 concerns the normalization of historically attested spellings and graphemic conventions toward modern forms through linguistically motivated substitutions.

The results provide qualified support for this hypothesis. While aggregate normalization rates are similar once long-s substitutions are accounted for (Table~\ref{tab:h3_examples}), the composition of normalization differs substantially. TrOCR’s normalization behavior is largely confined to graphemic conventions such as ligature decomposition, whereas Qwen additionally produces lexically motivated spelling regularizations that map historically valid forms to modern equivalents. This indicates that orthographic normalization in Vision--Language Models manifests as selective, linguistically driven regularization beyond what is warranted by the visual signal alone, with important implications for diplomatically faithful transcription.

\begin{table}[t]
\centering
\caption{Representative examples of orthographic normalization. Qwen exhibits lexically motivated normalization beyond graphemic conventions, whereas TrOCR’s normalization is largely limited to ligature handling.}
\label{tab:h3_examples}
\begin{tabular}{p{2.0cm} p{2.0cm} p{4.0cm} p{4.0cm}}
\hline
\textbf{Type} & \textbf{Model} & \textbf{Ground Truth} & \textbf{OCR Output} \\
\hline
Lexical & Qwen & Antient history & Ancient history \\
Lexical & Qwen & imploy their thoughts & employ their thoughts \\
Lexical & Qwen & phænomena & phaenomena \\
\hdashline
Ligature & Qwen & Celtæ & Celtae \\
Ligature & Qwen & Cæſar & Caeſar \\
Ligature & Qwen & Ætius & Aetius \\
\hdashline
Ligature & TrOCR & Celtæ & Celteze \\
Ligature & TrOCR & Cæſar & Caeſar \\
Ligature & TrOCR & Ætius & Aetius \\
\hline
\end{tabular}
\end{table}

\emph{Error Propagation Hypothesis (H4).}  
Hypothesis~H4 concerns differences in error propagation dynamics between architectures.

Empirical analysis shows that TrOCR exhibits larger and more contiguous deviation spans, indicating greater susceptibility to cascading errors once local alignment is disrupted. In contrast, Qwen more often produces bounded deviations that remain locally constrained, even when errors occur. As illustrated in Table~\ref{tab:h4_examples}, these differences reflect distinct error dynamics rather than differences in overall accuracy and have direct consequences for manual correction effort and editorial workflow design.

\begin{table}[t]
\centering
\caption{Representative examples illustrating error propagation and fragmentation (H4). Both models exhibit comparable edit distances, but differ in how errors spread within a line.}
\label{tab:h4_examples}
\begin{tabular}{p{1.4cm} p{5.4cm} p{5.4cm}}
\hline
\textbf{Model} & \textbf{Ground Truth} & \textbf{OCR Output} \\
\hline
Qwen &
Veſpaſian, i. 6. and ſucceeded by Julius Frontinus, &
and ſucceeded by Julius Frontinus, \\
\hdashline
Qwen &
the revolting barons are ſubdued by the prince a &
the revolted barons are ſubdued by the prince at \\
\hdashline
TrOCR &
—; lords of, ſee Lords. &
—jects of, ſee hards. \\
\hdashline
TrOCR &
his inequality to fulfil the conjugal duties; and he con- &
his incapable to fulfil the conjugal duties; and he con- \\
\hline
\end{tabular}
\end{table}

As shown in Section~4, TrOCR exhibits a steeper degradation with increasing sequence length, whereas Qwen’s errors remain more bounded. This difference reflects architectural sensitivity to sequence length and has direct implications for downstream correction effort on long or information-dense lines.

\subsection{Architectural Synthesis and Implications}

Taken together, the four hypotheses show that architectural design choices produce distinct and predictable error regimes with different downstream consequences.

TrOCR’s character-level supervision favors visually grounded substitutions that preserve glyph appearance but increase cascading deviation when alignment fails.

Qwen’s vision–language pretraining constrains orthographic drift through semantically guided decoding. Rather than cascading visual substitutions, it tends toward normalization and occasional semantic shifts, limiting structural deviation but increasing the risk of silent transformation.

Crucially, these behaviors represent architectural trade-offs rather than absolute strengths or weaknesses. While OCR-native models privilege visual fidelity at the cost of increased correction effort, Vision--Language Models constrain error growth at the cost of introducing normalization risk. Importantly, these risks are systematic and predictable rather than arbitrary. When they are explicitly understood and monitored, the substantially lower overall error rates achieved by Vision--Language Models may outweigh their disadvantages even in editorially demanding contexts.

In practice, model selection is a calibrated decision based on tolerance for different forms of risk. Vision–Language Models reduce error volume but introduce normalization risk, whereas OCR-native models preserve visual fidelity at the cost of increased correction effort.

Rather than prescribing a single optimal OCR architecture, we argue that model selection should follow a thorough error analysis that makes architectural risk profiles explicit. Users should choose models whose characteristic error behaviors align with their editorial goals, available expertise, and acceptable risk thresholds, rather than relying solely on aggregate accuracy metrics. This perspective motivates the broader synthesis in Section~6, where architectural bias is considered as a general property of OCR systems rather than a model-specific comparison.

\section{Discussion and Conclusions}

This study examined how architectural design choices shape OCR error behavior under historical domain shift. By focusing on error structure rather than aggregate accuracy, we showed that recognition systems with similar performance metrics can differ substantially in their reliability for scholarly use.

\subsection{Architectural Bias, Evaluation, and Editorial Practice}

Architectural inductive bias systematically shapes how models resolve visual uncertainty, producing distinct error regimes that differ in locality, detectability, and scholarly impact. These differences persist even when aggregate metrics are comparable, indicating that OCR systems are not interchangeable within historical digitization pipelines.

From a practical perspective, post-correction and editorial workflows must therefore be architecture-aware. Different error regimes require different safeguards, correction strategies, and levels of human oversight. Aligning evaluation and workflow design with known model behavior is essential for scalable and trustworthy historical digitization, particularly where cascading deviations or silent normalization carry scholarly risk.

\subsection{Limitations and Future Work}

This study isolates line-level recognition under controlled segmentation to focus on recognition behavior. Future work should extend this analysis to page-level pipelines where layout analysis, line detection, and recognition interact more strongly. Applying similar methods to non-Latin scripts, manuscript materials, and multilingual corpora would further test the generality of this perspective. Developing benchmarks that explicitly encode transcription intent and scholarly risk remains an important open challenge.

\subsection{Conclusion}

Architectural design fundamentally shapes OCR error behavior and associated scholarly risk. Model selection is therefore not only a technical decision but one that affects the reliability and interpretation of digitized historical texts. By linking error patterns to architectural properties, this work supports more transparent, use-aware evaluation and deployment of OCR systems.

\begin{credits}
\subsubsection{\ackname}This research was supported by funding from Oskar Öflunds Stiftelse and Jalmari Finnen Säätiö. The study is connected to the national DARIAH-FI research infrastructure, which promotes collaboration among Finnish research groups. We also thank the Helsinki Computational History Group (COMHIS) for their support and facilitation. Computational resources were provided by CSC – IT Center for Science under the ECCO RE-OCR project (ID 2005488).
\subsubsection{\discintname}
The authors have no competing interests to declare.
\end{credits}
\bibliographystyle{splncs04}
\bibliography{references}

\end{document}